\documentclass[preprint,groupedaddress, twocolumn, tightenlines, 10pt]{revtex4-2}
\usepackage{amsmath}

\usepackage[short]{optidef}

\usepackage{float}
\makeatletter
\let\newfloat\newfloat@ltx
\makeatother

\usepackage{listings}
\usepackage{hyperref}

\usepackage[draft]{minted} 

\makeatletter
\def\PYG@reset{\let\PYG@it=\relax \let\PYG@bf=\relax%
    \let\PYG@ul=\relax \let\PYG@tc=\relax%
    \let\PYG@bc=\relax \let\PYG@ff=\relax}
\def\PYG@tok#1{\csname PYG@tok@#1\endcsname}
\def\PYG@toks#1+{\ifx\relax#1\empty\else%
    \PYG@tok{#1}\expandafter\PYG@toks\fi}
\def\PYG@do#1{\PYG@bc{\PYG@tc{\PYG@ul{%
    \PYG@it{\PYG@bf{\PYG@ff{#1}}}}}}}
\def\PYG#1#2{\PYG@reset\PYG@toks#1+\relax+\PYG@do{#2}}

\@namedef{PYG@tok@w}{\def\PYG@tc##1{\textcolor[rgb]{0.73,0.73,0.73}{##1}}}
\@namedef{PYG@tok@c}{\let\PYG@it=\textit\def\PYG@tc##1{\textcolor[rgb]{0.24,0.48,0.48}{##1}}}
\@namedef{PYG@tok@cp}{\def\PYG@tc##1{\textcolor[rgb]{0.61,0.40,0.00}{##1}}}
\@namedef{PYG@tok@k}{\let\PYG@bf=\textbf\def\PYG@tc##1{\textcolor[rgb]{0.00,0.50,0.00}{##1}}}
\@namedef{PYG@tok@kp}{\def\PYG@tc##1{\textcolor[rgb]{0.00,0.50,0.00}{##1}}}
\@namedef{PYG@tok@kt}{\def\PYG@tc##1{\textcolor[rgb]{0.69,0.00,0.25}{##1}}}
\@namedef{PYG@tok@o}{\def\PYG@tc##1{\textcolor[rgb]{0.40,0.40,0.40}{##1}}}
\@namedef{PYG@tok@ow}{\let\PYG@bf=\textbf\def\PYG@tc##1{\textcolor[rgb]{0.67,0.13,1.00}{##1}}}
\@namedef{PYG@tok@nb}{\def\PYG@tc##1{\textcolor[rgb]{0.00,0.50,0.00}{##1}}}
\@namedef{PYG@tok@nf}{\def\PYG@tc##1{\textcolor[rgb]{0.00,0.00,1.00}{##1}}}
\@namedef{PYG@tok@nc}{\let\PYG@bf=\textbf\def\PYG@tc##1{\textcolor[rgb]{0.00,0.00,1.00}{##1}}}
\@namedef{PYG@tok@nn}{\let\PYG@bf=\textbf\def\PYG@tc##1{\textcolor[rgb]{0.00,0.00,1.00}{##1}}}
\@namedef{PYG@tok@ne}{\let\PYG@bf=\textbf\def\PYG@tc##1{\textcolor[rgb]{0.80,0.25,0.22}{##1}}}
\@namedef{PYG@tok@nv}{\def\PYG@tc##1{\textcolor[rgb]{0.10,0.09,0.49}{##1}}}
\@namedef{PYG@tok@no}{\def\PYG@tc##1{\textcolor[rgb]{0.53,0.00,0.00}{##1}}}
\@namedef{PYG@tok@nl}{\def\PYG@tc##1{\textcolor[rgb]{0.46,0.46,0.00}{##1}}}
\@namedef{PYG@tok@ni}{\let\PYG@bf=\textbf\def\PYG@tc##1{\textcolor[rgb]{0.44,0.44,0.44}{##1}}}
\@namedef{PYG@tok@na}{\def\PYG@tc##1{\textcolor[rgb]{0.41,0.47,0.13}{##1}}}
\@namedef{PYG@tok@nt}{\let\PYG@bf=\textbf\def\PYG@tc##1{\textcolor[rgb]{0.00,0.50,0.00}{##1}}}
\@namedef{PYG@tok@nd}{\def\PYG@tc##1{\textcolor[rgb]{0.67,0.13,1.00}{##1}}}
\@namedef{PYG@tok@s}{\def\PYG@tc##1{\textcolor[rgb]{0.73,0.13,0.13}{##1}}}
\@namedef{PYG@tok@sd}{\let\PYG@it=\textit\def\PYG@tc##1{\textcolor[rgb]{0.73,0.13,0.13}{##1}}}
\@namedef{PYG@tok@si}{\let\PYG@bf=\textbf\def\PYG@tc##1{\textcolor[rgb]{0.64,0.35,0.47}{##1}}}
\@namedef{PYG@tok@se}{\let\PYG@bf=\textbf\def\PYG@tc##1{\textcolor[rgb]{0.67,0.36,0.12}{##1}}}
\@namedef{PYG@tok@sr}{\def\PYG@tc##1{\textcolor[rgb]{0.64,0.35,0.47}{##1}}}
\@namedef{PYG@tok@ss}{\def\PYG@tc##1{\textcolor[rgb]{0.10,0.09,0.49}{##1}}}
\@namedef{PYG@tok@sx}{\def\PYG@tc##1{\textcolor[rgb]{0.00,0.50,0.00}{##1}}}
\@namedef{PYG@tok@m}{\def\PYG@tc##1{\textcolor[rgb]{0.40,0.40,0.40}{##1}}}
\@namedef{PYG@tok@gh}{\let\PYG@bf=\textbf\def\PYG@tc##1{\textcolor[rgb]{0.00,0.00,0.50}{##1}}}
\@namedef{PYG@tok@gu}{\let\PYG@bf=\textbf\def\PYG@tc##1{\textcolor[rgb]{0.50,0.00,0.50}{##1}}}
\@namedef{PYG@tok@gd}{\def\PYG@tc##1{\textcolor[rgb]{0.63,0.00,0.00}{##1}}}
\@namedef{PYG@tok@gi}{\def\PYG@tc##1{\textcolor[rgb]{0.00,0.52,0.00}{##1}}}
\@namedef{PYG@tok@gr}{\def\PYG@tc##1{\textcolor[rgb]{0.89,0.00,0.00}{##1}}}
\@namedef{PYG@tok@ge}{\let\PYG@it=\textit}
\@namedef{PYG@tok@gs}{\let\PYG@bf=\textbf}
\@namedef{PYG@tok@gp}{\let\PYG@bf=\textbf\def\PYG@tc##1{\textcolor[rgb]{0.00,0.00,0.50}{##1}}}
\@namedef{PYG@tok@go}{\def\PYG@tc##1{\textcolor[rgb]{0.44,0.44,0.44}{##1}}}
\@namedef{PYG@tok@gt}{\def\PYG@tc##1{\textcolor[rgb]{0.00,0.27,0.87}{##1}}}
\@namedef{PYG@tok@err}{\def\PYG@bc##1{{\setlength{\fboxsep}{\string -\fboxrule}\fcolorbox[rgb]{1.00,0.00,0.00}{1,1,1}{\strut ##1}}}}
\@namedef{PYG@tok@kc}{\let\PYG@bf=\textbf\def\PYG@tc##1{\textcolor[rgb]{0.00,0.50,0.00}{##1}}}
\@namedef{PYG@tok@kd}{\let\PYG@bf=\textbf\def\PYG@tc##1{\textcolor[rgb]{0.00,0.50,0.00}{##1}}}
\@namedef{PYG@tok@kn}{\let\PYG@bf=\textbf\def\PYG@tc##1{\textcolor[rgb]{0.00,0.50,0.00}{##1}}}
\@namedef{PYG@tok@kr}{\let\PYG@bf=\textbf\def\PYG@tc##1{\textcolor[rgb]{0.00,0.50,0.00}{##1}}}
\@namedef{PYG@tok@bp}{\def\PYG@tc##1{\textcolor[rgb]{0.00,0.50,0.00}{##1}}}
\@namedef{PYG@tok@fm}{\def\PYG@tc##1{\textcolor[rgb]{0.00,0.00,1.00}{##1}}}
\@namedef{PYG@tok@vc}{\def\PYG@tc##1{\textcolor[rgb]{0.10,0.09,0.49}{##1}}}
\@namedef{PYG@tok@vg}{\def\PYG@tc##1{\textcolor[rgb]{0.10,0.09,0.49}{##1}}}
\@namedef{PYG@tok@vi}{\def\PYG@tc##1{\textcolor[rgb]{0.10,0.09,0.49}{##1}}}
\@namedef{PYG@tok@vm}{\def\PYG@tc##1{\textcolor[rgb]{0.10,0.09,0.49}{##1}}}
\@namedef{PYG@tok@sa}{\def\PYG@tc##1{\textcolor[rgb]{0.73,0.13,0.13}{##1}}}
\@namedef{PYG@tok@sb}{\def\PYG@tc##1{\textcolor[rgb]{0.73,0.13,0.13}{##1}}}
\@namedef{PYG@tok@sc}{\def\PYG@tc##1{\textcolor[rgb]{0.73,0.13,0.13}{##1}}}
\@namedef{PYG@tok@dl}{\def\PYG@tc##1{\textcolor[rgb]{0.73,0.13,0.13}{##1}}}
\@namedef{PYG@tok@s2}{\def\PYG@tc##1{\textcolor[rgb]{0.73,0.13,0.13}{##1}}}
\@namedef{PYG@tok@sh}{\def\PYG@tc##1{\textcolor[rgb]{0.73,0.13,0.13}{##1}}}
\@namedef{PYG@tok@s1}{\def\PYG@tc##1{\textcolor[rgb]{0.73,0.13,0.13}{##1}}}
\@namedef{PYG@tok@mb}{\def\PYG@tc##1{\textcolor[rgb]{0.40,0.40,0.40}{##1}}}
\@namedef{PYG@tok@mf}{\def\PYG@tc##1{\textcolor[rgb]{0.40,0.40,0.40}{##1}}}
\@namedef{PYG@tok@mh}{\def\PYG@tc##1{\textcolor[rgb]{0.40,0.40,0.40}{##1}}}
\@namedef{PYG@tok@mi}{\def\PYG@tc##1{\textcolor[rgb]{0.40,0.40,0.40}{##1}}}
\@namedef{PYG@tok@il}{\def\PYG@tc##1{\textcolor[rgb]{0.40,0.40,0.40}{##1}}}
\@namedef{PYG@tok@mo}{\def\PYG@tc##1{\textcolor[rgb]{0.40,0.40,0.40}{##1}}}
\@namedef{PYG@tok@ch}{\let\PYG@it=\textit\def\PYG@tc##1{\textcolor[rgb]{0.24,0.48,0.48}{##1}}}
\@namedef{PYG@tok@cm}{\let\PYG@it=\textit\def\PYG@tc##1{\textcolor[rgb]{0.24,0.48,0.48}{##1}}}
\@namedef{PYG@tok@cpf}{\let\PYG@it=\textit\def\PYG@tc##1{\textcolor[rgb]{0.24,0.48,0.48}{##1}}}
\@namedef{PYG@tok@c1}{\let\PYG@it=\textit\def\PYG@tc##1{\textcolor[rgb]{0.24,0.48,0.48}{##1}}}
\@namedef{PYG@tok@cs}{\let\PYG@it=\textit\def\PYG@tc##1{\textcolor[rgb]{0.24,0.48,0.48}{##1}}}

\def\PYGZus{\char`\_}

\makeatother

\usepackage{algorithm2e}
\RestyleAlgo{ruled}
\SetAlgoCaptionLayout{raggedright}

\begin{document}


\title{JAXFit: Trust Region Method for Nonlinear Least-Squares Curve Fitting on the GPU}


\author{Lucas R. Hofer}\email{lucas.hofer@physics.ox.ac.uk}
\author{Milan Krstaji\'{c}}
\author{Robert P. Smith}
\affiliation{Clarendon Laboratory, University of Oxford, Parks Road, Oxford OX1 3PU, United Kingdom}


\date{\today}

\begin{abstract}
We implement a trust region method on the GPU for nonlinear least squares curve fitting problems using a new deep learning Python library called JAX. Our open source package, JAXFit, works for both unconstrained and constrained curve fitting problems and allows the fit functions to be defined in Python alone---without any specialized knowledge of either the GPU or CUDA programming. Since JAXFit runs on the GPU, it is much faster than CPU based libraries and even other GPU based libraries, despite being very easy to use. Additionally, due to JAX's deep learning foundations, the Jacobian in JAXFit's trust region algorithm is calculated with automatic differentiation, rather than than using derivative approximations or requiring the user to define the fit function's partial derivatives.
\end{abstract}


\maketitle

\section{Introduction}

Nonlinear least squares regression (NLSQ), colloquially known as curve fitting, is an essential tool in many fields and is used to determine the set of parameters which cause a model function to best match observational data. If the model function represents the data well, the extracted parameters often allow useful information to be recovered from the observed data. For example, images of laser beams are often fit with 2D Gaussians to measure the beams' size \cite{Siegman98, hoferlaser} and in biomedicine NLSQ regression is used to determine the location of cells in microscopy images  \cite{bates2008super}.

Unlike linear least squares regression, NLSQ regression cannot be solved analytically and numerical methods must be used instead. Algorithms for performing NLSQ regression have existed for centuries with Gauss improving on Newton's method to give the Gauss-Newton method \cite{gauss}. Levenberg \cite{levenberg}---and later Marquardt \cite{marquardt}---added a damping term to the Gauss-Newton method to form the Levenberg-Marquardt method.  By reformulating the Levenberg-Marquardt algorithm as a trust region problem \cite{more2}, Mor\'e \cite{more1} developed a robust NLSQ algorithm which is still used today. Although the aforementioned methods address NLSQ regression problems without parameter constraints, Coleman and Li \cite{coleman1} developed a constrained NLSQ regression method, which they further expanded for large data \cite{coleman2}.

The advent of modern computing has greatly enhanced the utility of these NLSQ regression algorithms and there are several well established software packages such as MATLAB's \texttt{curvefit toolbox} \cite{matlab}, MINPACK \cite{minpack} and SciPy \texttt{optimize} \cite{scipy} which have have been robustly tested. These packages rely on the central processing unit (CPU) for all of the computation and when regressing a relatively small number of data points (e.g. $<10^4$) these algorithms run quickly; however, regressing a large number of data points, which is quite common when fitting image data, can drastically increase the fit time. The advent of graphical processing units (GPUs)---which can parallelize operations on their many cores---means these NLSQ regression algorithms can be significantly sped up as most of the computation time is spent on parallelizable operations such as matrix multiplication.

In the past, programming the GPU required specialized knowledge, usually of CUDA, which is likely why only a few frameworks implement NLSQ regression algorithms on the GPU \cite{gaunt, lmufit}. Even Gpufit \cite{gpufit}, which allows its Levenberg-Marquardt algorithm to be called from MATLAB or Python, requires the user to define both custom fit functions as well as the function's partial derivatives with CUDA before rebuilding the entire project with a C++ compiler.

In this paper we introduce our open source package JAXFit \cite{jaxfit}, which implements both unconstrained and constrained NLSQ regression algorithms on the GPU and solely in Python by using a relatively new deep learning package called JAX \cite{jax_original}. JAX executes Python defined operations on the GPU and additionally has in-built automatic differentiation of functions \cite{jax_software}, which removes the need to explicitly define the fit function's partial derivatives. JAXFit is easy to use---working as a drop in replacement for SciPy's curve fit function---and achieves significant speed-ups due to the GPU.

The rest of the paper is laid out as follows: in Section~\ref{sec:trm} we detail the trust region method algorithm. For readers interested in the implementation alone, we recommend skipping to Section~\ref{sec:implementation} in which the speed-up of the trust region method using JAX is described. Finally, in Section~\ref{sec:freespeed} we evaluate the fitting speed of JAXFit in comparison with both SciPy and Gpufit. 

\section{Trust Region Method Algorithm}\label{sec:trm}
 
We use SciPy's trust region method algorithm---which is an amalgamation of both Mor\'e's \cite{more1,more2} and others' \cite{numeric, coleman1} work---as the basis for our own trust region method (TRM) implementation. SciPy is the default library for curve fitting in Python and since, to our knowledge, the SciPy TRM algorithm has been only partially explained in a blog post \cite{mayorov_trf} by the author, we describe the unconstrained TRM algorithm below. JAXFit also implements SciPy's adapted version of the Coleman-Li \cite{coleman1, coleman2} method for constrained NLSQ regression, but we neglect to detail it.

We first define a function $h(y; \mathbf{x})$ which relies on an independent variable $y$ and $n$ fixed parameters $\left\{x_1, x_2, ..., x_n\right\}$ which comprises a parameter vector $\mathbf{x}$. If we have $v$ number of observations $\left\{y_i, z_i\right\}$ which we want to model with $h$ we first define a residual function $r_i$
\begin{equation}
r_i(x) = h(y_i, \mathbf{x}) - z_i\text{.}
\label{eq:residual}
\end{equation}
\noindent Now, to find the parameters $\mathbf{x}$ which best approximate the observed data, the least squares of the residual functions are minimized as follows
\begin{mini}
{\mathbf{x}}{f(\mathbf{x}) = \frac{1}{2}\sum_i^v r_i^2(\mathbf{x})=\frac{1}{2}\mathbf{r}^T(\mathbf{x})\mathbf{r}(\mathbf{x})\text{.}}{}{}
\label{eq:msqe}
\end{mini}
\noindent where $\mathbf{r}(\mathbf{x})$ is a vector containing the $v$ residual functions. For linear functions Eq.~\ref{eq:msqe} can be analytically solved; however, for nonlinear functions numerical methods must be used and these generally involve iterative steps in which the parameter vector $\mathbf{x}$ is successively updated with a parameter update vector $\mathbf{w}$. 

The problem then is to find the parameter update vector $\mathbf{w}$ which minimizes $f(\mathbf{x}+\mathbf{w})$. To do this, a quadratic model of $f(\mathbf{x}+\mathbf{w})$ is formed using Taylor's theorem \cite{numeric},
\begin{equation}
f(\mathbf{x}+\mathbf{w})\approx f(\mathbf{x}) + \nabla f(\mathbf{x})^T\mathbf{w} +\frac{1}{2}\mathbf{w}^TH(\mathbf{x})\mathbf{w}\text{,}
\label{eq:taylor}
\end{equation}
\noindent where the expansion has been truncated after the quadratic term. The gradient $\nabla f(\mathbf{x})$ is given by  
\begin{equation}
\nabla f(\mathbf{x})=\sum_{i=1}^v\nabla r_i(\mathbf{x})r_i(\mathbf{x})=J(\mathbf{x})^T\mathbf{r}(\mathbf{x})
\label{eq:gradient}
\end{equation}
\noindent where the Jacobian $J(\mathbf{x})$ is
\[
J(\mathbf{x}) =
\begin{bmatrix}
  \frac{\partial r_1}{\partial x_1} & \frac{\partial r_1}{\partial x_2}& \dots &  \frac{\partial r_1}{\partial x_n} \\[1ex] 
  \frac{\partial r_2}{\partial x_1} & \frac{\partial r_2}{\partial x_2}& \dots &  \frac{\partial r_2}{\partial x_n} \\[1ex]
   \vdots & \vdots  & \ddots & \vdots \\[1ex]
  \frac{\partial r_v}{\partial x_1} &  \frac{\partial r_v}{\partial x_2}& \dots & \frac{\partial r_v}{\partial x_n}
\end{bmatrix}\text{.}\label{eq:jacobian}
\]
\noindent The Hessian, $H(\mathbf{x})$, in Eq.~\ref{eq:taylor} is given by
\begin{equation}
H(\mathbf{x})=\sum_{i=1}^v\nabla r_i(\mathbf{x})\nabla r_i(\mathbf{x})^T + \sum_{i=1}^v r_i(\mathbf{x})\nabla^2 r_i(\mathbf{x})\text{;}
\label{eq:hessian}
\end{equation}
\noindent however, since either the Laplacians or residuals in the second term of Eq.~\ref{eq:hessian} are often quite small, it is common in NLSQ algorithms to ignore it and approximate the Hessian with only the first term \cite{numeric}. Additionally, the first term can be rewritten using the Jacobian such that the approximate Hessian is $H(\mathbf{x})\approx J^T(\mathbf{x})J(\mathbf{x})$. Finally Eq.~\ref{eq:taylor} is rewritten such that $\mathbf{x}$ is implicit in the various terms
\begin{equation}
m(\mathbf{w}) = f + \mathbf{g}^T\mathbf{w} +\frac{1}{2}\mathbf{w}^TB\mathbf{w}\text{,}
\label{eq:mp}
\end{equation}
\noindent where $m(\mathbf{w})\approx f(\mathbf{x}+\mathbf{w})$, $f=f(\mathbf{x})$ is the function, $\mathbf{g}=\nabla f(\mathbf{x})$ is the gradient and $B=J^T(\mathbf{x})J(\mathbf{x})$ is the approximate Hessian.

\begin{figure}[t!]
\centering
  \includegraphics[scale=0.7]{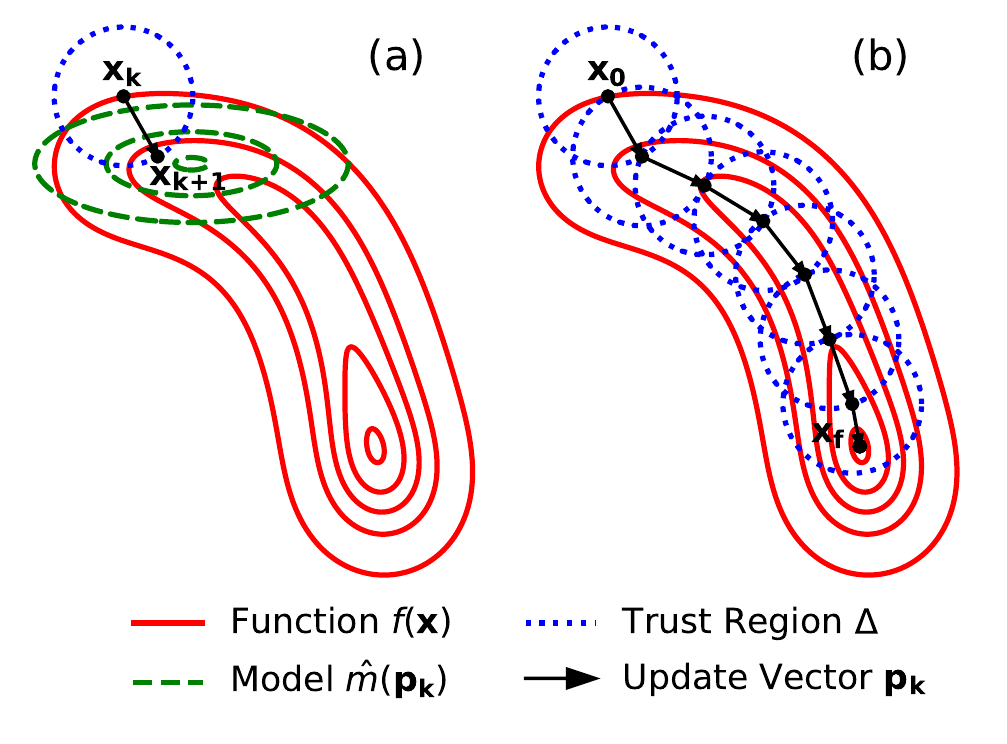}
  \caption{\label{fig:contour}Visualization of the trust region method algorithm. The parameter vector $\mathbf{x}$ is iteratively updated to find the minimum of the function $f(\mathbf{x})$ (solid contour lines, see Eq.~\ref{eq:msqe}). (a) A single update step $k$ in which an approximate quadratic model $\hat{m}(\mathbf{p_k})$ (dashed contour lines) of the function is minimized within a trust region of radius $\Delta$ (dotted line) to determine the parameter update vector $\mathbf{p_k}$ (see Algorithm~\ref{alg:one}). (b) Multiple update steps---with their respective trust regions---showing the convergence of the algorithm from the seed parameter vector $\mathbf{x_0}$ to the final parameter vector $\mathbf{x_f}$ at the minimum of $f(\mathbf{x})$. Although these trust region sizes are all the same, they will often differ in practice.}
\end{figure}

A parameter update $\mathbf{w}$ is now needed such that the quadratic model in Eq.~\ref{eq:mp} is minimized. Since, the model is only good as a local approximation, a trust region radius $\Delta$ is defined within the parameter space such that
\begin{mini}
  {\mathbf{w}}{m(\mathbf{w}) = f + \mathbf{g}^T\mathbf{w} +\frac{1}{2}\mathbf{w}^TB\mathbf{w}}{}{}
  \addConstraint{\left\lVert D\mathbf{w}\right\rVert\leq\Delta}{}
\label{eq:min}
\end{mini}
\noindent where $D$ is a---usually diagonal---scaling matrix used to make the trust region an ellipsoid rather than a sphere as the different parameters $\left\{x_1, x_2, ..., x_n\right\}$ comprising $\mathbf{x}$ can have vastly different scalings. However, according to Mor\'e \cite{more2}, this is equivalent to solving the problem in a scaled parameter space where the trust region is a sphere (see Fig.~\ref{fig:contour}a) and the scaled update vector is $\mathbf{p} = D\mathbf{w}$. In this case Eq.~\ref{eq:min} is equivalent to
\begin{mini}
  {\mathbf{p}}{\hat{m}(\mathbf{p})=f + \mathbf{\hat{g}}^T\mathbf{p} +\frac{1}{2}\mathbf{p}^T\hat{B}\mathbf{p}}{}{}
  \addConstraint{\left\lVert \mathbf{p}\right\rVert\leq\Delta}{}
\text{.}
\label{eq:hat}
\end{mini}
\noindent where $\mathbf{\hat{g}}=\hat{J}^T \mathbf{r}$, $\hat{B}= \hat{J}^T(\mathbf{x})\hat{J}(\mathbf{x})$ and $\hat{J}(\mathbf{x})=J(\mathbf{x})D^{-1}$ with the~$\hat{}$~symbol denoting the scaled space.

Mor\'e and Sorensen \cite{more2} showed that the parameter update $\mathbf{p}$ which is a solution to Eq.~\ref{eq:hat} is also the solution to the following 
\begin{align}
\left(\hat{B} + \alpha I\right)\mathbf{p} & = -\mathbf{\hat{g}} \label{eq:lm}\\
\alpha\left(\Delta- \left\lVert \mathbf{p}\right\rVert\right)& = 0  \label{eq:pbound}\\
\left(\hat{B} + \alpha I\right) & \text{ is positive semidefinite}  \nonumber
\end{align}
\noindent where $I$ is the identity matrix and the Levenberg-Marquardt parameter is a scalar $\alpha\geq0$. This is the problem we solve.

A basic outline of the TRM algorithm (see Algorithm~\ref{alg:one}) can now be given. After being seeded with an initial parameter vector $\mathbf{x_0}$ and initial trust radius $\Delta_0$, the parameter vector is iteratively updated for each step $k$ using Eq.~\ref{eq:lm}-\ref{eq:pbound} until a given accuracy is reached (see Fig.~\ref{fig:contour}b). A trial update step with $\alpha=0$ is first attempted for which $\mathbf{p_t}= -\hat{B}^{-1}\mathbf{\hat{g}}$ and if the trial update vector is within the trust region ($\left\lVert \mathbf{p_t}\right\rVert\leq\Delta$) then it is accepted as the parameter update vector. However, if it lies outside the trust region then the trial parameter update vector must be discarded as in actuality $\alpha>0$. Furthermore, in this case, the Levenberg-Marquardt parameter $\alpha$ must then be explicitly solved for as a sub-problem.

To solve for $\alpha$, we follow Mor\'e's \cite{more1} numerical method in which an initial guess $\alpha_0=0$ is iteratively updated with $q$ steps  (see Algorithm~\ref{alg:alpha}). Since, according to Eq.~\ref{eq:pbound}, the parameter update vector lies on the trust region radius ($\left\lVert \mathbf{p}\right\rVert - \Delta=0$) when $\alpha > 0$, we iteratively update $\alpha_q$ until $\mathbf{p}$ lies within a distance $\sigma\Delta$ of the trust radius---where $\sigma$ is the numerical accuracy. Formally this is $\left|\phi(\alpha) \right| \leq \sigma\Delta$ where $\phi(\alpha_q)=\left\lVert \mathbf{p}\right\rVert - \Delta$. Solving Eq.~\ref{eq:lm} for $\mathbf{p}$ allows $\phi(\alpha_q)$ to be rewritten as
\begin{equation}
\phi(\alpha_q) = \left\lVert\left(\hat{B}+\alpha_q I\right)^{-1}\mathbf{\hat{g}}\right\rVert - \Delta\text{.}
\label{eq:phi}
\end{equation}
\noindent Using Eq.~\ref{eq:phi} and $\phi ' \left(\alpha_q\right)$, the derivative of $\phi\left(\alpha_q\right)$ with respect to $\alpha_q$, the update to the Levenberg-Marquardt parameter is given by
\begin{equation}
\alpha_{q+1}=\alpha_q - \left(\frac{\phi\left(\alpha_q\right)+\Delta}{\Delta}\right)\left(\frac{\phi\left(\alpha_q\right)}{\phi ' \left(\alpha_q\right)}\right)
\label{eq:alphaupdate}
\end{equation}

\setlength{\textfloatsep}{10pt}
\begin{algorithm}[h!]
\caption{Overall trust region method algorithm. The Levenberg-Marquardt (LM) parameter and update sub-problems are detailed in Algorithm~\ref{alg:alpha} and Algorithm~\ref{alg:trr} respectively.}\label{alg:one}
$\mathbf{x_k} \gets \mathbf{x_0}$\;
$\Delta_k \gets \Delta_0$\;
\While{not accuracy condition}
{
	$\mathbf{p_t} \gets -\hat{B}^{-1}\mathbf{\hat{g}}$\;
	\eIf{$\left\lVert \mathbf{p_t} \right\rVert \leq\Delta_k$}
	{
   	 	$\mathbf{p_k} \gets \mathbf{p_t}$\;
	}
	{
   	 	$\alpha_k \gets$ LM parameter, Algorithm~\ref{alg:alpha}\;
		$\mathbf{p_k} \gets \left(\hat{B} + \alpha_k I\right)^{-1}\mathbf{\hat{g}}$\;
   	 }
	$\mathbf{x_{k+1}}$, $\Delta_{k+1} \gets$ Update, Algorithm~\ref{alg:trr} $(\Delta_k, \mathbf{x_k}, \mathbf{p_k})$\;
}
\end{algorithm}

\begin{algorithm}[h!]
\caption{Solving for the Levenberg-Marquardt parameter $\alpha$ when the update vector $\mathbf{p}$ lies on the trust region radius.}\label{alg:alpha}
$\alpha_q \gets \alpha_0$\;
$u_q \gets \frac{\left\lVert \mathbf{\hat{g}}\right\rVert}{\Delta}$\;
$l_q \gets -\frac{\phi\left(\alpha_0\right)}{\phi ' \left(\alpha_0\right)}$\;
\While{$\left|\phi(\alpha_q)\right| \leq \sigma\Delta$}
{
	$\alpha_t \gets \alpha_q - \left(\frac{\phi\left(\alpha_q\right)+\Delta}{\Delta}\right)\left(\frac{\phi\left(\alpha_q\right)}{\phi 		' \left(\alpha_q\right)}\right)$\;
	\eIf{$l_q\leq \alpha_t \leq u_q$}
	{
   	 	$\alpha_{q+1} \gets \alpha_t$\;
	}
	{
   	 	 $\alpha_{q+1} \gets \text{max}\left\{0.001 u_q, \left(l_qu_q\right)^\frac{1}{2}\right\}$\;
   	 }
	$l_{q+1} \gets \text{max}\left\{l_q, \alpha_q - \frac{\phi\left(\alpha_q\right)}{\phi ' \left(\alpha_q\right)}\right\}$\;
	\eIf{$\phi\left(\alpha_q\right)< 0$}
	{
   	 	$u_{q+1} \gets \alpha_q$\;
	}
	{
   	 	 $u_{q+1} \gets u_q$\;
   	 }
}
\end{algorithm}

\begin{algorithm}[h!]
\caption{Determining the updated parameter vector $\mathbf{x_{k+1}}$ and trust region radius $\Delta_{k+1}$.}\label{alg:trr}
	$\mathbf{w_k} \gets D^{-1}\mathbf{p_k}$\;
	\uIf{$\gamma(\mathbf{x_k}, \mathbf{w_k}) > 0.75$}{
   	 	$\mathbf{x_{k+1}} \gets \mathbf{x_k} + \mathbf{w_k}$\;
		$\Delta_{k+1} \gets 2\Delta_k$\;
  	}
	\uElseIf{$0.25 \leq \gamma(\mathbf{x_k}, \mathbf{w_k}) \leq 0.75$}
	{
    		$\mathbf{x_{k+1}} \gets \mathbf{x_k}+\mathbf{w_k}$ \;
		$\Delta_{k+1} \gets \Delta_k$\;
  	}
  	\Else{
    		 $\mathbf{x_{k+1}} \gets \mathbf{x_k}$\;
		 $\Delta_{k+1} \gets 0.25\left\lVert \mathbf{p_k}\right\rVert $;
  	}
\end{algorithm}

\noindent subect to the condition that $l_q\leq \alpha_{q+1}\leq u_q$, where $l_q$ and $u_q$ are the lower and upper bounds \cite{more1} respectively (see Algorithm~\ref{alg:alpha}). If the update lies outside these bounds, it is instead given by 
\begin{equation}
\alpha_{q+1}=\text{max}\left\{0.001 u_q, \left(l_qu_q\right)^\frac{1}{2}\right\}\text{.}
\end{equation}
\noindent When given a reasonable initial value for $\alpha_0$ this method for determining the Levenberg-Marquardt parameter generally converges to a solution within a few iterations. This solution, in conjunction with Eq.~\ref{eq:lm}, is then used to calculate the scaled update vector $\mathbf{p}$ and thereafter the unscaled parameter update vector $\mathbf{w}=D^{-1}\mathbf{p}$.

Once the parameter update vector has been found, it is important to determine how much it improves the accuracy and whether $\mathbf{w}$ should be accepted as an update to the parameter vector $\mathbf{x}$. To do this, the following equation is used \cite{numeric}
\begin{equation}
\gamma(\mathbf{x}, \mathbf{w})=\frac{f(\mathbf{x}) - f(\mathbf{x}+\mathbf{w})}{m(0) - m(\mathbf{w})}
\label{eq:accuracy}
\end{equation}
\noindent which gives the ratio of the actual reduction over the predicted reduction. If $\gamma(\mathbf{x}, \mathbf{w})$ is larger than a set threshold value, then the update $\mathbf{w}$ is accepted and the trust region radius is either kept the same or increased (see Algorithm~\ref{alg:trr}). However, if $\gamma(\mathbf{x}, \mathbf{w})$ is lower than the threshold value then the parameter vector is kept the same and the size of the trust region radius is decreased.

\section{Implementing the Trust Region Algorithm with JAX}\label{sec:implementation}

Our goal when designing JAXFit was to have a fast, yet easy to use NLSQ regression algorithm. We therefore decided to use Python as it is a popular data analysis language featuring several libraries which implement array operations on the GPU. Python's popularity as a data analysis language is due in large part to open source packages such as NumPy \cite{numpy}, SciPy \cite{scipy} and Pandas \cite{mckinney2011pandas}. NumPy provides efficient routines for array operations on the CPU, including an extensive set of linear algebra functions, whereas SciPy complements NumPy with a large set of scientific functions and algorithms. 

Although SciPy features NLSQ fitting, its original---and currently default---implementation simply wraps MINPACK's \cite{minpack} Fortran Levenberg-Marquardt algorithm for unconstrained NLSQ problems. Nikolay Mayorov \cite{mayorov_trf} significantly improved SciPy's curve fitting code with a more modern approach by writing a trust region algorithm for both unconstrained (based on More \cite{more1, more2}, see Section~\ref{sec:trm}) and constrained NLSQ problems \cite{coleman1, coleman1} solely in Python. 

Unlike NumPy for CPU array operations, a similar unifying Python library for array operations on the GPU has not been established. We decided to use a relative newcomer, JAX, for GPU array operations---which since its release by Google in 2018 has seen rapid adoption by both the machine learning and scientific computing communities. JAX describes itself as ``NumPy on the GPU" and it implements almost all of NumPy's functionality and a significant portion of SciPy's on the GPU. 

In addition to running on the GPU, JAX also speeds up operations through use of Just in Time (JIT) compiliation in which JIT wrapped Python functions are compiled to optimized machine code at run time. When coupled with a GPU this massively speeds-up execution times as JAX converts the JIT wrapped functions to highly optimized Accelerated Linear Algebra (XLA) code \cite{jax_original} which runs efficiently on the GPU hardware. Thus, JAX allows code to be written in a high-level language (Python), but with the performance of a hardware specific language (XLA).

Although for large array operations JAX is significantly faster than NumPy, it is often slower for small array operations due to the overhead of transferring data to the GPU. Therefore, we only move large array operations (array sizes $\geq v$, where $v$ is the number of data points being fit) to JAX and the GPU, while leaving the small array operations to NumPy and the CPU. The operations we run with JAX include calculating the function $f(\mathbf{x})$, the gradient $\nabla f(\mathbf{x})$, the quadratic model $m(\mathbf{w})$ and the loss function.

We also to rely on JAX to calculate the Jacobian $J(\mathbf{x})$ and to perform singular value decomposition (SVD, see Appendix~\ref{sec:svd}) of the Jacobian matrix---which is a robust method for solving the linear least squares problem in Eq.~\ref{eq:lm} \footnote{Even though SciPy uses a LSMR algorithm \cite{lsmr} for solving the least squares problem when the Jacobian matrix is both large and sparse, JAX does not currently support most SciPy sparse matrix operations and we therefore implement only SVD.}. When calculating the Jacobian, we use JAX's in-built automatic differentiation \cite{autodiff} rather than either requiring the user to provide partial derivatives (as in Gpufit) or using approximations to the partial derivatives (the SciPy default). 

Even though JAX is both fast and easy to use, its dynamism comes at a price. When creating optimized machine code, JAX traces out a computational graph of the function using abstract tracer arrays for the function inputs---which can be computationally expensive. However, the computational overhead pays off when the code is repeatedly called. 

The functional style of programming used in SciPy's TRM code caused significant retracing to occur and to avoid this we incorporated the TRM algorithm into a Python object. The object's JAX functions are traced only when first called and successive calls to the object's curve fit function run without tracing. Additionally, since changing the size of a JIT function's input array also causes retracing to occur, we developed an array zero-padding method which fixes the size of the JIT functions' input arrays (see Appendix~\ref{sec:atrace}). 

\section{Fit Performance Comparison}\label{sec:freespeed}

\begin{figure}[t!]
\centering
  \includegraphics[scale=0.7]{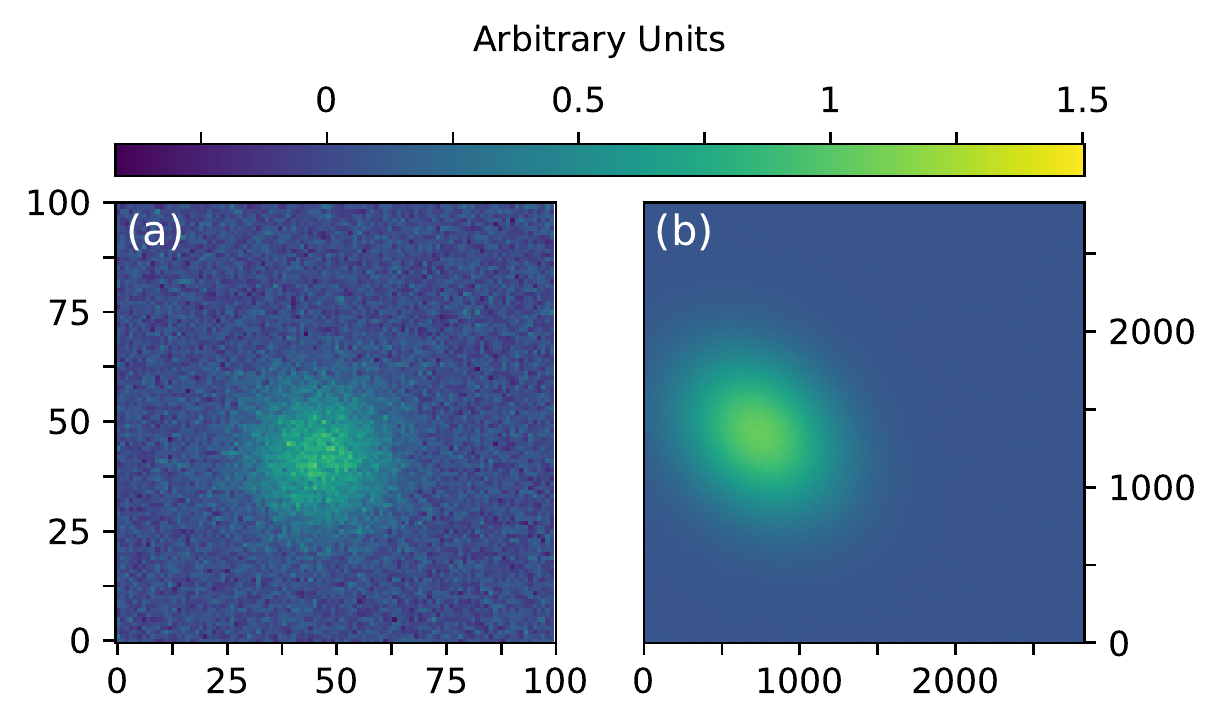} 
  \caption{\label{fig:sample} Dataset sample images. (a) Image with the smallest data length, $v=10^4$. (b)  Image with the largest data length, $v=8\times10^6$. Although the simulated Gaussian noise in (b) has the same standard deviation as the noise in (a), it is visually smoothed out due to the large number of data points.}
\end{figure}

We compared the fit speed of JAXFit against both SciPy and Gpufit by creating a simulated dataset. The dataset is comprised of 2D rotated, elliptical Gaussians (see Figure~\ref{fig:sample}) which feature seven fitting parameters \cite{hofer2017scale}. We chose fifteen data lengths (number of residuals $v$) evenly distributed between $v=10^4$ and $v=8\times10^6$ on a logarithmic scale and simulated fifty-one 2D Gaussian images for each data length. The seven fitting parameters for each 2D Gaussian image were drawn from a uniform, random distribution. Gaussian distributed noise was also added to each image with a mean of zero and a fixed standard deviation for all the images in the dataset.

\setlength{\textfloatsep}{20pt plus 2pt minus 4pt} 

\begin{figure}
\centering
  \includegraphics[scale=0.7]{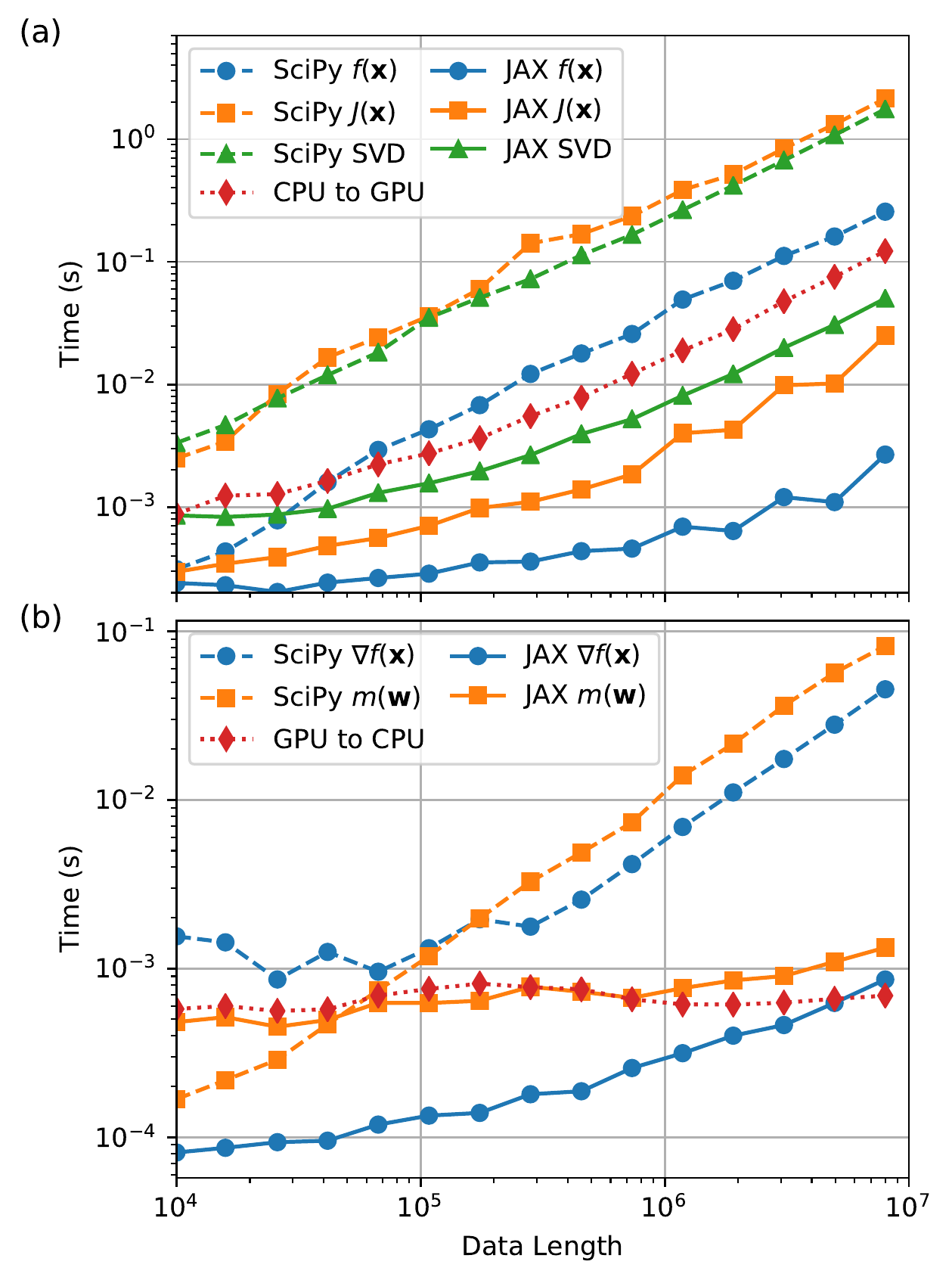}
  \caption{\label{fig:scipy_jax} Comparison of SciPy vs. JAXFit (JAX) for individual operations. (a) Speed of the more computationally expensive operations---including function evaluation $f(\mathbf{x})$, Jacobian calculation $J(\mathbf{x})$ and singular value decomposition (SVD) of the Jacobian---as a function of the data length. Additionally, the time it takes to transfer fit data from the CPU to the GPU is shown. (b) Other operations, including the gradient calculation $\nabla f(\mathbf{x})$ and evaluation of the quadratic model $m(\mathbf{w})$ as a function of the data length. The summative CPU to GPU array transfer time for one iteration step within the trust region method algorithm is also shown.}
\end{figure}

JAXFit and SciPy use the same TRM algorithm which enables us to compare the speed of the large array operations---which run on the GPU and CPU for JAXFit and SciPy respectively. The dataset's simulated Gaussian images were fit for each data length in turn using both JAXFit and SciPy. We then averaged the fit speed for each of the fifteen data lengths (see Fig.~\ref{fig:scipy_jax}) to show the operation speed vs. the data length; however, for each data length the first fit (tracing) was not included in the time average as tracing can generally be avoided (see Appendix~\ref{sec:atrace}).

As expected, the JAXFit operations-----including the function evaluations $f(\mathbf{x})$, calculation of Jacobians $J(\mathbf{x})$, SVD of the Jacobian matrix, calculation of the gradient $\nabla f(\mathbf{x})$ and evaluation of the model function $m(\mathbf{w})$---were significantly faster than SciPy's. Although the JAXFit operations generally scaled linearly with the data length, a few of the operations including SVD and the gradient calculation showed a flat region for small data lengths. This is likely due to not all the GPU resources being utilized. However, since the CPU resources are generally fully utilized, almost all the SciPy operations scale linearly---irrespective of the data length.

\begin{figure}
\centering
  \includegraphics[scale=0.7]{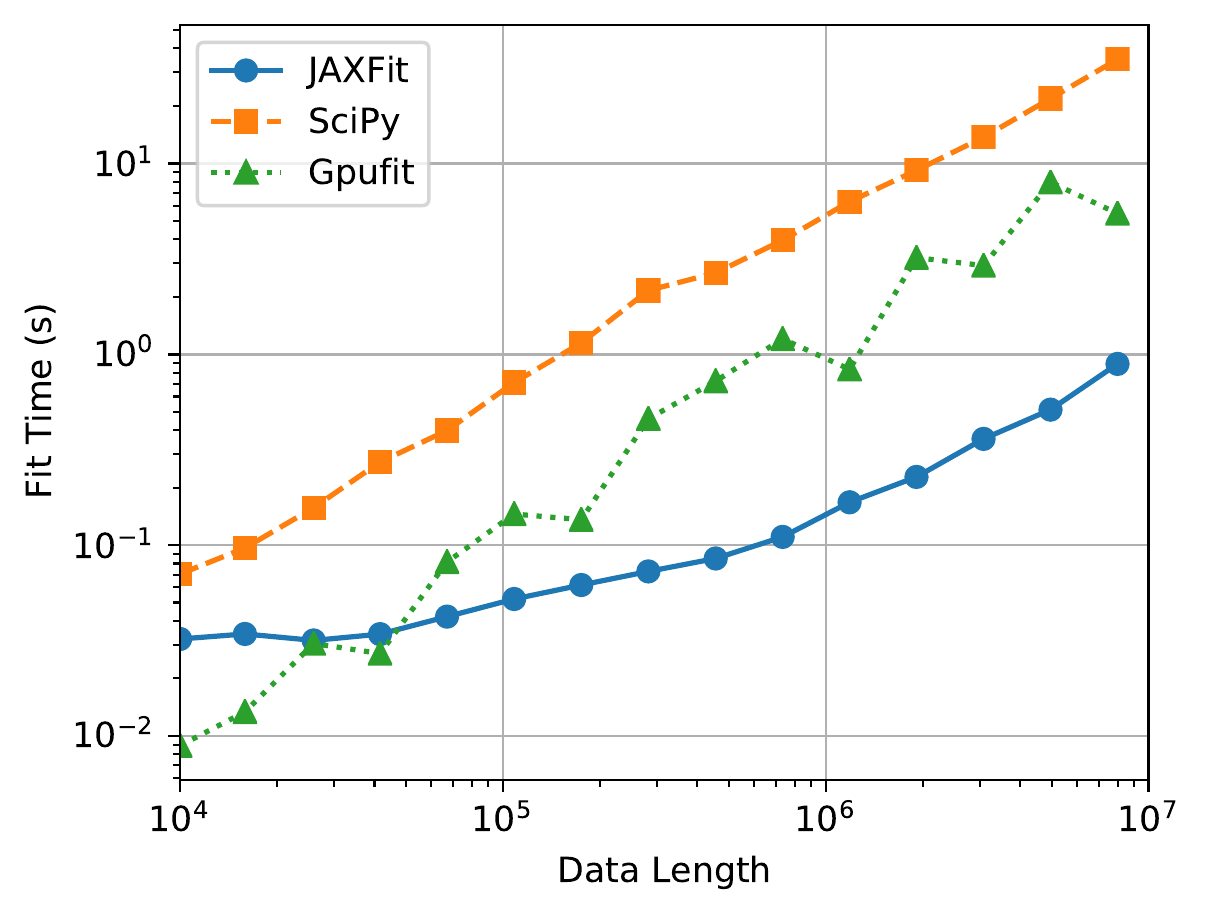}
  \caption{\label{fig:comparison}Overall fit speed of JAXFit, SciPy and Gpufit as a function of the data length. JAXFit (solid-circle) and Gpufit (dotted-triangle) are much faster than SciPy (dash-square) as they utilize the GPU rather than CPU. Although Gpufit is slightly faster for a small data lengths ($v<3\times 10^4$), JAXFit is significantly faster for large data lengths.}
\end{figure}

In addition to the computation time, we measured the time required to move the fit data from the CPU to the GPU (see Fig.~\ref{fig:scipy_jax}a), as well as the time to transfer data back from the GPU to CPU (see Fig.~\ref{fig:scipy_jax}b). As expected, the CPU to GPU time scales linearly with the size of the data being transferred, whereas the GPU to CPU transfer time is roughly constant---since the size of the arrays transferred back are solely dependent on the number of fitting parameters $n$.

We also perform an overall fit speed comparison of JAXFit, SciPy and Gpufit for unconstrained NLSQ regression (see Fig.~\ref{fig:comparison}). Using each method we measured the overall fit time for all the simulated Gaussian images in the dataset. We again ignored the first image for each data length and averaged the remaining fifty fit times. As expected, JAXFit and Gpufit are much faster than SciPy due to the GPU vs. CPU hardware advantage. 

When comparing the GPU methods, Gpufit is slightly faster than JAXFit if the data length is small (${v<3\times 10^4}$), as its Levenberg-Marquardt algorithm runs entirely in CUDA on the GPU. However, as the data length increases, this advantage quickly disappears and JAXFit significantly outperforms Gpufit when the data length is large---almost an order of magnitude faster for $v>10^6$. Since the number of iterations it takes the algorithms in JAXFit and Gpufit to converge is roughly equal when using the same fit data, JAXFit's better performance is likely due to the highly optimized XLA library underpinning JAX, which uses the GPU hardware resources more efficiently.

Finally, we note that these experiments were all run in a Google Colab notebook featuring a single 2 GHz Intel Xeon CPU core and a NVIDIA Tesla V100-SXM2-16GB GPU---which is already several years old and is roughly equivalent to NVIDIA's latest series of gaming GPUs (RTX3080/RTX3090). Although the speed of JAXFit will depend primarily on the GPU specifications, even when using a lower spec'd GPU, JAXFit is much faster than CPU based NLSQ regression methods.

\section{Conclusion}

We have implemented a trust region method for both unconstrained and constrained nonlinear least squares (NLSQ) curve fitting on the GPU using a deep learning Python library, JAX. Our open source package, JAXFit \cite{jaxfit}, allows the user to dynamically define fit functions in Python and, unlike other GPU NLSQ libraries, requires no specialized knowledge of CUDA or the GPU. Furthermore, the partial derivatives which comprise the Jacobian in the trust region algorithm are calculated with automatic differentiation---rather than either requiring the user to define the partial derivatives or using approximations.

JAXFit significantly outperforms SciPy's CPU based NSLQ curve fitting function due to its use of the GPU. Furthermore, JAXFit is easier to use and generally faster than another GPU fitting library (Gpufit) since it makes more efficient use of the GPU resources. However, JAXFit's simplicity comes with the computational overhead of JAX's function tracing. Although the initial trace is unavoidable, JAXFit is generally able to avoid retracing even when the size of the fit data changes (see Appendix~\ref{sec:atrace}) and thus remains very fast.

For ease of use, JAXFit is designed as a drop-in replacement for SciPy's curve fit function. Additionally, JAXFit incorporates features not previously described such as handling fit data uncertainty (see Appendix~\ref{sec:uncertainty}), scaling of the Jacobian and implementation of various loss functions---see the Github repository for all the available features \cite{jaxfit}.

\appendix

\section{Avoiding Retracing of Dynamic Array Sizes}\label{sec:atrace}

JAX needs to retrace a JIT function whenever the size of the input array changes. Although this isn't a problem for neural networks---which it was primarily constructed for---it can be an issue for curve fitting. For example in our laboratory, we fit images of cold atom clouds \cite{hoferatom} with 2D Gaussians or 2D parabolas and since the size of the clouds can be very different, we often need to fit different sized data arrays. 

To circumvent the JAX retracing process for these different sized arrays, JAXFit allows the user to define a fixed input data size $s$.  When the fit data passed to JAXFit is smaller than this fixed size then the fit data is padded with dummy data to account for the difference between the actual data length $v$ and the fixed data length $s$. 


%
%
%

\begin{listing}[b!]

\begin{Verbatim}[commandchars=\\\{\}]
\PYG{k+kn}{import} \PYG{n+nn}{jax.numpy} \PYG{k}{as} \PYG{n+nn}{jnp}
\PYG{k+kn}{import} \PYG{n+nn}{jax.jit} \PYG{k}{as} \PYG{n+nn}{jit}

\PYG{n+nd}{@jit}
\PYG{k}{def} \PYG{n+nf}{masked\PYGZus{}jac}\PYG{p}{(}\PYG{n}{coords}\PYG{p}{,} \PYG{n}{args}\PYG{p}{,} \PYG{n}{data\PYGZus{}mask}\PYG{p}{)}
     \PYG{n}{J} \PYG{o}{=} \PYG{n}{jac\PYGZus{}func}\PYG{p}{(}\PYG{n}{coords}\PYG{p}{,} \PYG{n}{args}\PYG{p}{)}
     \PYG{n}{J\PYGZus{}masked} \PYG{o}{=} \PYG{n}{jnp}\PYG{o}{.}\PYG{n}{where}\PYG{p}{(}\PYG{n}{data\PYGZus{}mask}\PYG{p}{,} \PYG{n}{J}\PYG{p}{,} \PYG{l+m+mi}{0}\PYG{p}{)}
     \PYG{k}{return} \PYG{n}{J\PYGZus{}masked}
\end{Verbatim}

\caption{Example of masking dummy data within a JIT wrapped Python function.}
\label{lst:jac}
\end{listing}

This does require the user to pick a sensible value for $s$. If $s\gg v$, then significant time is wasted processing the dummy data. On the other hand, if the actual data size is larger than the fixed input data size ($v>s$) then the JAXFit functions must be retraced. Thus, when all the data being analyzed is roughly the same size, the optimal fixed size is $s=v_{\text{max}}$ where $v_{\text{max}}$ is the largest data size being fit. However, if the fit data sizes differ significantly then it can be more efficient to instantiate multiple curve fit objects---each with a different fixed $s$ to handle the range of data sizes.

Lastly, simply padding the input fit data with dummy data would lead to incorrect fits. Thus, we've added masking operations (see Listing~\ref{lst:jac}) throughout the TRM algorithm so that the dummy data doesn't affect the fit, but merely operates as a placeholder to keep JAX from retracing the functions.

\section{Solving Linear Least Squares Equation with Singular Value Decomposition}\label{sec:svd}

Singular value decomposition \cite{martin2012extraordinary} factors any matrix $M$ which has size $v\times n$ into the product of three matrices 
\begin{equation}
M=U\Sigma V^T
\end{equation}
\noindent where $U$ is a $v\times v$ orthonormal matrix, $\Sigma$ is a $v\times n$ diagonal matrix and $V$ is a $n\times n$ orthonormal matrix. Since both $U$ and $V$ are orthonormal $U^TU=UU^T=I$ (with the same expression for $V$) where $I$ is the identity matrix.

The special properties of SVD allow the least squares problem (see Eq.~\ref{eq:lm}) in the trust region method algorithm to be solved. We start by rewriting Eq.~\ref{eq:lm} explicitly in terms of the scaled Jacobian $J$ as
\begin{equation}
\left(J^TJ + \alpha I\right)\mathbf{p} = -J^T\mathbf{r}\text{,}
\label{eq:jaclm}
\end{equation}
\noindent where the scaled space's hat notation $\hat{}$ has been dropped. Now using SVD, the Jacobian is $J=U\Sigma V^T$ and its transpose is  $J^T=V\Sigma^T U^T$.  Multiplying these two together and using the orthonormality of the matrix $U$ gives $J^TJ=V\Sigma^T\Sigma V^T$. Plugging all this into Eq.~\ref{eq:jaclm}
\begin{equation}
\left(V\Sigma^T\Sigma V^T+ \alpha I\right)\mathbf{p}=V\Sigma^T U^T\mathbf{r}
\end{equation}
\noindent and multiplying from the left on both sides of this equation by $V^T$ yields
\begin{equation}
\left(\Sigma^T\Sigma+ \alpha I\right) V^T \mathbf{p}=\Sigma^T U^T\mathbf{r}\text{.}
\label{eq:wh}
\end{equation}
\noindent Finally, solving Eq.~\ref{eq:wh} for $\mathbf{p}$ gives
\begin{equation}
\mathbf{p} = V \left(\Sigma^T\Sigma+ \alpha I\right)^{-1}\Sigma^T U^T\mathbf{r}
\label{eq:reducedlm}
\end{equation}
\noindent which is significantly simpler to work with as $\Sigma^T\Sigma+ \alpha I$ is diagonal and therefore easily invertible. 

The SVD of the scaled Jacobian is done once per parameter update step $k$ (see Algorithm~\ref{alg:one}), as is calculating both $\Sigma^T\Sigma$ and $\Sigma^T U^T\mathbf{r}$. Even though Eq.~\ref{eq:reducedlm} may be called multiple times within a single update step $k$ when solving for the Levenberg-Marquardt parameter (see Algorithm~\ref{alg:alpha}), the matrices involved in this repeated calculation are all of size $n\times n$ and the operation is therefore computationally inexpensive.

\section{Dealing with Uncertainty}\label{sec:uncertainty}

Sometimes the data points $\left\{y_i, z_i\right\}$ which are being fit with the function $h(y; \mathbf{x})$ have different weightings/errors $\sigma_i$ which need to be taken into consideration. These errors modifiy Eq.~\ref{eq:msqe} to the following form \cite{gavin}
\begin{equation}
f(\mathbf{x}) = \frac{1}{2}\sum_i^v \left[\frac{h(y_i, \mathbf{x}) - z_i}{\sigma_i}\right]^2=\frac{1}{2}\mathbf{r}^T(\mathbf{x})W\mathbf{r}(\mathbf{x})
\label{eq:explicit}
\end{equation}
\noindent where the residual vector is explicitly defined using Eq.~\ref{eq:residual} and where $W$ is the weighting matrix given by
\begin{equation}
W =
\begin{bmatrix}
  \frac{1}{\sigma_{11}^2} & 0 & \dots &  0 \\[1ex] 
  0 &  \frac{1}{\sigma_{22}^2} & \dots &  0 \\[1ex]
   \vdots & \vdots  & \ddots & \vdots \\[1ex]
  0 &  0& \dots & \frac{1}{\sigma_{vv}^2}
\end{bmatrix}\text{.}\label{eq:weighting}
\end{equation}
\noindent However, this diagonal weighting matrix is a special case and the more general form is $W=C^{-1}$ \cite{gavin} where $C$ is the symmetric covariance matrix given by 
\begin{equation}
C =
\begin{bmatrix}
  \sigma_{11}^2 & \sigma_{12}^2 & \dots &  \sigma_{1v}^2 \\[1ex] 
  \sigma_{21}^2 & \sigma_{22}^2 & \dots &  \sigma_{2v}^2 \\[1ex]
   \vdots & \vdots  & \ddots & \vdots \\[1ex]
  \sigma_{v1}^2 &  \sigma_{v2}^2 & \dots &\sigma_{vv}^2
\end{bmatrix}\text{.}\label{eq:covariance}
\end{equation}
By using Eq.~\ref{eq:explicit} instead of Eq.~\ref{eq:msqe} and working through the math in Section~\ref{sec:trm} the gradient (Eq.~\ref{eq:gradient}) and approximated Hessian are modified to be 
\begin{equation}
\mathbf{g}=\nabla f(\mathbf{x})=J(\mathbf{x})^TW\mathbf{r}(\mathbf{x})
\end{equation}
\noindent and
\begin{equation}
H(\mathbf{x})\approx B= J^T(\mathbf{x})WJ(\mathbf{x})\text{.}
\end{equation}
\noindent respectively. This causes Eq.~\ref{eq:lm} to have the following form
\begin{equation}
\left[J^TWJ + \alpha I\right]\mathbf{p} = -J^TW\mathbf{r},
\label{eq:wlm}
\end{equation}

\noindent where, again, the scaled space's hat notation $\hat{}$ has been removed for readability. The weighting matrix in Eq.~\ref{eq:wlm} seemingly complicates the algorithmic implementation in Section~\ref{sec:trm}; however, we show below that Eq.~\ref{eq:wlm} can be rewritten in the same form as Eq.~\ref{eq:lm} thereby keeping the algorithm the same.

Since by definition the covariance matrix $C$ is real, Hermitian and positive-definite it can be factorized via Cholesky decomposition as $C=LL^T$ where $L$ is a real, lower triangular matrix. The weighting matrix can then be defined in terms of this factorization $W=C^{-1}=\left(L^T\right)^{-1}L^{-1}=\left(L^{-1}\right)^TL^{-1}$ and substituted into Eq.~\ref{eq:wlm}
\begin{equation}
\left[J^T\left(L^{-1}\right)^TL^{-1}J + \alpha I\right]\mathbf{p} = -J^T\left(L^{-1}\right)^TL^{-1}\mathbf{r}\text{.}
\label{eq:wlm2}
\end{equation}
\noindent By using matrix transpose and multiplication rules Eq.~\ref{eq:wlm2} is rewritten as
\begin{equation}
\left[\tilde{J}^T\tilde{J} + \alpha I\right]\mathbf{p} = -\tilde{J}^T\tilde{\mathbf{r}}\text{.}
\label{eq:tlm}
\end{equation}
\noindent where $\tilde{J}=L^{-1}J$ and $\tilde{\mathbf{r}}=L^{-1}\mathbf{r}$---which successfully gives us the same form as Eq.~\ref{eq:lm}. 

Thus, when the user gives a covariance matrix, the weighted residual vector $\tilde{\mathbf{r}}$ and weighted Jacobian $\tilde{J}$ are determined before following the algorithm in Section~\ref{sec:trm}. To do this, the Cholesky decomposition is first performed to find $L$. Since calculating the inverse of $L$ can be computationally expensive the following two equations are solved to find $\tilde{J}$ and $\tilde{\mathbf{r}}$
\begin{align}
L\tilde{J}&=J\label{eq:slsq1}\\
L\tilde{\mathbf{r}}&=\mathbf{r}\text{.}\label{eq:slsq2}
\end{align}

However, it is rare to use the full covariance matrix and often only the diagonal elements are known---which results in the weighting matrix given by Eq.~\ref{eq:weighting}. In this case the inverted form of $L$ is easily found to be
\begin{equation}
L^{-1} =
\begin{bmatrix}
  \frac{1}{\sigma_{11}} & 0 & \dots &  0 \\[1ex] 
  0 &  \frac{1}{\sigma_{22}} & \dots &  0 \\[1ex]
   \vdots & \vdots  & \ddots & \vdots \\[1ex]
  0 &  0& \dots & \frac{1}{\sigma_{vv}}
\end{bmatrix}\text{.}\label{eq:easesig}
\end{equation}
\noindent which means $\tilde{J}$ and $\tilde{\mathbf{r}}$ can be calculated without having to solve Eqs.~\ref{eq:slsq1}-\ref{eq:slsq2}. Additionally, this allows vector rather than matrix multiplication to be used
\begin{align}
\tilde{J}&=L^{-1}J=\left(\mathbf{S}^{-1}\right)^TJ\label{eq:easy_jac}\\
\tilde{\mathbf{r}}&=L^{-1}\mathbf{r}=\left(\mathbf{S}^{-1}\right)^T\mathbf{r}\label{eq:easy_res}
\end{align}
\noindent which is a less computationally expensive operation. $\mathbf{S}$ is a vector containing the diagonal elements of $L$
\begin{equation}
\mathbf{S} =
\begin{bmatrix}
  \sigma_{11} \\ 
  \sigma_{22} \\
  \vdots \\
  \sigma_{vv}  
\end{bmatrix}\text{.}\label{eq:svector}
\end{equation}

Similar to SciPy's curve fit, JAXFit allows the user to give either $\mathbf{S}$ or $C$ for the fit data error when calling the curve fit function. At the beginning of the algorithm $\left(\mathbf{S}^{-1}\right)^T$ is calculated if given $\mathbf{S}$, whereas if $C$ is given it is factorized via Cholesky decomposition to yield $L$. During the iterative part of the algorithm (see Algorithm~\ref{alg:one}), the weighted residual vector $\tilde{\mathbf{r}}$ must be determined for each parameter update step $k$ using either Eq.~\ref{eq:slsq2} and $\left(\mathbf{S}^{-1}\right)^T$  or Eq.~\ref{eq:easy_res} and $L$---depending on whether $\mathbf{S}$ or $C$ has been given.

The weighted Jacobian $\tilde{J}$ also needs to be calculated for each parameter update step $k$. By default JAXFit calculates the Jacobian with automatic differentiation and $\tilde{J}$ is therefore calculated by simply passing the weighted residual function $\tilde{\mathbf{r}}$ to JAX's \texttt{autodiff} function. However, JAXFit also allows users to explicitly define the Jacobian matrix---in terms of $h(y; \mathbf{x})$---in which case $\tilde{J}$ must be solved for explicitly using either Eq.~\ref{eq:slsq1} or Eq.~\ref{eq:easy_jac}.

\vskip .2cm
\section*{Funding}
This work was supported by EPSRC Grant Nos. EP/P009565/1 and EP/TO19913/1, the John Fell Oxford University
Press (OUP) Research Fund and the Royal Society.

\vskip .2cm
\section*{Acknowledgements}
L.R. thanks Oliver Karnbach and Mark Ijspeert for helpful discussions. 

\vskip .2cm
\section*{Disclosures}

The authors declare no conflicts of interest. After finishing this paper, we became aware of a recently released Levenberg-Marquardt algorithm using JAX \cite{blondel2021}.

\vskip .2cm
\section*{Data Availability}

We make code available in \cite{jaxfit}. 

\vskip .2cm
\bibliography{refs}

\end{document}